\documentclass[conference,compsoc]{IEEEtran}
\IEEEoverridecommandlockouts
\usepackage{color}
\usepackage{ulem}
\ifCLASSOPTIONcompsoc
  \usepackage[nocompress]{cite}
\else
  \usepackage{cite}
\fi
\usepackage[pdftex]{graphicx}
\graphicspath{{./}}
\DeclareGraphicsExtensions{.pdf,.jpeg,.png}

\ifCLASSINFOpdf
\else
\fi
\usepackage{amsmath}
\begin{document}
%
\title{Physical Artificial Intelligence: The Concept Expansion of Next-Generation Artificial Intelligence}

\makeatletter
\newcommand{\linebreakand}{%
  \end{@IEEEauthorhalign}
  \hfill\mbox{}\par
  \mbox{}\hfill\begin{@IEEEauthorhalign}
}
\makeatother

\author{\IEEEauthorblockN{Yingbo Li}
\IEEEauthorblockA{
Hainan University\\
lantuzi@aliyun.com}

\and
\IEEEauthorblockN{Yucong Duan* \thanks{*Corresponding author: duanyucong@hotmail.com}}
\IEEEauthorblockA{
Hainan University\\
duanyucong@hotmail.com}

\and
\IEEEauthorblockN{Anamaria-Beatrice Spulber}
\IEEEauthorblockA{
Visionogy\\
anne@visionogy.com}

\and
\IEEEauthorblockN{Haoyang Che }
\IEEEauthorblockA{
Zeekr Group,\\
haoyang.che@zeekrlife.com}

\linebreakand
\and
\IEEEauthorblockN{Zakaria Maamar}
\IEEEauthorblockA{
Zayed University\\
zakaria.maamar@zu.ac.ae}

\and
\IEEEauthorblockN{Zhao Li}
\IEEEauthorblockA{
Alibaba Group \\
lizhao.lz@alibaba-inc.com
}

\and
\IEEEauthorblockN{Chen Yang}
\IEEEauthorblockA{
Ghent University\\
Chen.Yang@UGent.be}

\and
\IEEEauthorblockN{Yu lei }
\IEEEauthorblockA{
Inner Mongolia University,\\
yuleiimu@sohu.com}
}


%


\maketitle

\begin{abstract}
Artificial Intelligence has been a growth catalyst to our society and is cosidered across all idustries as a fundamental technology. However, its development has been limited to the signal processing domain that relies on the generated and collected data from other sensors. In recent research, concepts of Digital Artificial Intelligence and Physicial Artifical Intelligence have emerged and this can be considered a big step in the theoretical development of Artifical Intelligence. In this paper we explore the concept of Physicial Artifical Intelligence and propose two subdomains: Integrated Physicial Artifical Intelligence and Distributed Physicial Artifical Intelligence. The paper will also examine the trend and governance of Physicial Artifical Intelligence.
\end{abstract}


%
\IEEEpeerreviewmaketitle

\section{Introduction}
Artificial Intelligence~(AI) has been one of the most popular topics in the Information and Communication Technologies~(ICT) field. AI powered the development of many advanced systems such as robotics. AI used to be confined to digital signal processing such as text processing, image object recognition, and speech recognition. However, when considering computer science holistically, signal processing is only a small part of the field. The AI applications have been extended to include robots, Internet of Things~(IoT), smart cities, ~etc. In~\cite{pai}, Miriyev and Kovac classified AI into Digital~AI which processes signals, and Physical~AI including physical robots. In this paper we explore the concept of Physical~AI and extend it to Integrated Physical~AI such as robots or Distributed Physical~AI such as IoT. In~\cite{pai}, authors considered Integrated Physical~AI as Physical AI, whose components are together and in a restricted space. We propose Distributed Physical~AI as a kind of Physical~AI too, whose components could be distributed in a wide space. The aalysis of the Physical AI concepts brings the opportunity to discuss about AI and Physical AI from a larger perspective. Additionally, it enables us to explore further the manifestations of Physical AI.

In this paper, we will begin by reviewing the state of the art of Artificial Intelligence, and conclude with a discussion about the concept of Physical Artificial Intelligence and how can we leverage the benefits of it across diffenrent domains. Throughout the paper we will review the trends in Physical Artificial Intelligence and the potential governence problem implications. Addtionally, we propose to use Knowledge Graph and Data-Information-Knowledge-Wisdom (DIKW) to further develop the research on Physical Artificial Intelligence. The intenrion of this paper is to advance the theoretical development of Physical Artificial Intelligence. 

\section{Overview of Artificial Intelligence}
AI is well known for its outperforming human capabilities in popular benchmarks such ImageNet~\cite{imagenet}. Its various industrial  applications, in both its own domains such as Natural Language Processing~(NLP), speech recognition, face detection, and image classification, or other disciplines such as agriculture,  biology, and chemistry has widely been recognized.

AI originates from the principle of building a Turing machine from neurons, a concept proposed by McCulloch and Pitts in 1943~\cite{mpnm}. Since 1988 different NN milestones such as Backpropagation algorithm continued to develop~\cite{backp}. Lecun invented Convolutional~Neural~Network~(CNN) with backpropagation in 1998, and in 2006 the fast training of Neural~Network~(NN) was addressed. Considering the above, both NN and even AI, have begun their fast pace development in 2012 ~\cite{hist}. To succeed, AI needs the support of advanced and affordable computing hardware such as GPU cards, and machine learning algorithms, especially~NN. The relationships between NN, AI, and other related concepts are illustrated in Figure~\ref{fig:ai}. NN is essential in powering the AI, however, the development of the AI~domain, involves various disciplines such as knowledge modelling, a highly researched discipline that involves Knowledge Graph to DIKW. With the increasing and fundamental importance of NN in mind, we will start by reviewing the history and successful algorithms of~NN.

The success of deep learning originates from deep NN, especially CNN applied to image classification~\cite{alom}. Primarily, most NN~algorithms were of supervised type, such as CNN and Recurrent~NN~(RNN). CNN and its variants are involved with the classification and recognition purpose such as image classifciation and face recognition. RNN is different from CNN as it considers the temporal information in the NN, and as such RNN, including its variant Long Short-Term Memory~(LSTM), has become popular in speech recognition and language translation uses. The semi-supervised learning such as Generative Adversarial Networks~(GAN) is often used in image generation, image enhancement, and video game~\cite{creswell}.

\begin{figure}[ht]
\centering
\includegraphics[width=1.0\columnwidth]{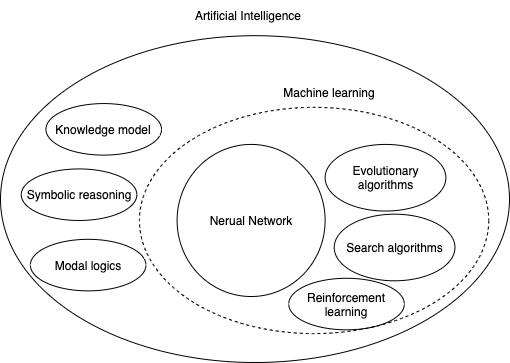}
\caption{Disciplines and techniques associated with AI \sout{AI architecture}}
\label{fig:ai}
\end{figure}

Deep learning algorithms could be classified into supervised, semi-supervised, unsupervised, and reinforcement learning based on supervision level during  the  model training period~\cite{mla}. At first, the deep learning most supervised algorithms have been extensively used in face recognition, text sentiment classification, speech recognition, and other similar cases. When the training data is not entirely labelled, variants of supervised deep learning algorithms such as semi-supervised learning algorithms could be used. The unsupervised learning on the other hand, does not rely on training data labelling but it learns from the internal relations caused by the initial defined features, such as Auto-Encoders(AE), GAN, and Restricted Boltzmann Machines~(RBM). In reinforcement  learning,  the  algorithms  can  only  obtain the  incremental  data instead of all pre-existing  data  in  each  processing step.    

Apart from computer science applications, AI has been used in academia and various industries. For example, it has been used to faciliate the prediction of the process of catalysis utilization \cite{catalysis}. Other uses involved the financial market, where AI has been used in the dynamic pricing and fraud detection \cite{fraud}. In the energy domain, AI is used to reduce the electricity \cite{elec} and solar modelling \cite{solar}. In the agriculture AI has been used in the detection of fruit ripening~\cite{fruit}.

Although AI has proved to be useful in various domains of research and industries, AI has also encountered a few limitations. Most of the current AI~applications are limited to the individual applications. One example is that CNN is useful in image classifiation and text classification, while RNN is useful in machine translation and speech recognition. AI still encounters challenges in mananging trivial details and annoying business rules, and some of these problems have been the focus of researchers \cite{compl}. Almost all AI algorithms need to understand binary codes or numbers, they lack of high logical  inference and  problem  solving capabilities that humans have, and this is mainly because not every real problem can be converted into pure mathematic problems. For example, AI finds it is hard to understand the sentence differences between the "Macbook can be used as the chopping board" and “Macbook is a computer” in the architecutre concept of DIKW~\cite{dikw}. In addition, AI mostly worked until now like a black, and while researchers know AI works well, they are not clear about the reasons behind its success for any specific problems. Therefore, Explainable Artificial Intelligence (XAI) \cite{xai} has been a research domain that is focused on discovering the reasons behind the success of some specific NN algorithms.

\section{The concept expansion of Physical Artificial Intelligence}
Currently the concept of Artificial Intelligence, as described in the above section, is related to processing the data and signals in the computer system. Even the hardware that is related to the AI only captures the input data and deliver the output data from the AI system, as illustrated in Figure~\ref{fig:hai}. One example is the Smart Home~\cite{smarth} supported by Amazon's Alexa speech assistant~\cite{alexa}. In~\cite{pai}, Miriyev and Kovac proposed the concept of Digital Artificial Intelligence~(DIAI) that refers to the current popular data-based and date-processing~AI.

\begin{figure}[ht]
\centering
\includegraphics[width=1.0\columnwidth]{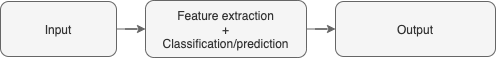}
\caption{The hardware architecture related to AI}\label{fig:hai}
\end{figure}

Contrary to Digital~AI, Miriyev and Kovac \cite{pai} have proposed the concept of Physical Artificial Intelligence (PAI), which refers to the nature-like robots that are driven by the intelligence. Miriyev and Kovac used the bee robot to explain the concept of PAI, a multi-discipline that combines autonomous robots, materials, structures, and perceptions. PAI requires the balance between software - the intelligent system, and hardware - material, mechanics, and etc. As illustrated in Figure~\ref{fig:pai}, PAI bears its roots in the materials science, mechanical engineering, computer science, chemistry and biology.

\begin{figure}[ht]
\centering
\includegraphics[width=0.6\columnwidth]{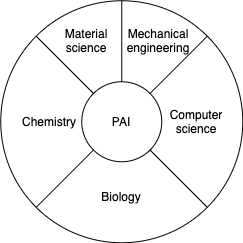}
\caption{Multidisciplinary nature of PAI}\label{fig:pai}
\end{figure}

In the proposed concept of PAI by Miriyev and Mirko~\cite{pai}\cite{spai}, PAI refers to the typical robot and robot system. In this paper we propose to extend the concept of PAI to all the potential applications identifying the advantages of AI for both the hardware and software. Several examples are used to explain the extended concept of PAI:

\begin{itemize}
\item PAI in IoT. IoT is the typical mixed application of the cloud, sensor, software and data analytics~\cite{iot}. The robot concatenates the hardware and software in one complete intelligent machine, while IoT can be distributed to either a small space such as a room or to a wide area such as the city. Since AI can be used to improve the stability of each node of the IoT such as a sensor or the central data analytics and predication, IoT is a fertile application domain for PAI. The node of IoT used for sensing and controlling needs the support of the science and technologies, materials, chemistry, mechanism, and computer science, even the biology. 

\item PAI in automobile. The self-driving car can be considered as a variant of the intelligent robot system. The self-driving car has the same necessary features as the normal robot: the sensor, the embedded computing module, the mechanical system, the new material and so on. The self-driving car is often connected to the Internet for navigation and the latter provides the IoT feature to the self-driving car.

\item PAI in agriculture. The agriculture is one of the most successful applications of Physical Artificial Intelligence. The sensors including the cameras, temperature meter and hygrometer are used to monitor the growth progress of the plants and predict the best harvest time. The defects are often detected to alarm the potential risk for a intervention. 

\item PAI in healthcare. The healthcare, especially the healthcare for the prevention, is a typical usage of Physical Artificial Intelligence. The biological sensors and the chemical sensors are used to monitor the old man and the patient to predict the potential risks such as falling or an unstable situation; the centrual center is notified by the edge device when the risk happens. The computing happends both at the edge sides and the centrual servers.

\item PAI in Logistics. PAI has been extensively used in multiple aspects of the logistics. The "last mile" is the expensive and hard problem of the logistic industry at involves parcel and food delivery. Some delivery robots and drones \cite{delivery} have been used in the delivery market to replace the humans. The automatic sorting robot has been used in the sorting center of the logistics \cite{logi}.  
\end{itemize}

In the above survey, the extended concept of PAI has been extensively used in multiple industries outside the robot industry. The concept of PAI is based on the interdiscipline research of five disciplines proposed in Figure \ref{fig:pai} \cite{pai}.

\section{The Trend of Physical Artificial Intelligence}

Until 2012 Digital Artificial Intelligence~(DIAI) mimicked the brain capability of logical thinking and induction in human brain, to process the data and signals percepted by human eyes and ears. As far as we know, the capabilities of human beings are not limited to the logical thinking of the brain. The brain of the human beings is only responsible for processing the signals and transmitting commands to other parts of the body, that are responsible for many functions, such as movement, vision perception, sound perception, digestion and etc. Therefore, DIAI just uncovers a limited part of the powerful potentials of AI, while PAI like a whole human body with respect to the whole human body, would heavily extend the application of AI from the academics to the industries.

PAI has the potential to use deep learning to mimick not only the individual human but also the human society as a whole. Robots are a typical example of Integrated~PAI~(IPAI) that mimick the individual humans, and integrates the perception of the physical world through multiple sensors that collect signals and data, the induction from multiple indices, and the physical response in the physical world as shown in Fig.~\ref{fig:ipai}, that illustrates the most important modules in IPAI. A robot's perception, computing, and mechanical modules are confined into a limited space, while similar to the human society Distributed~PAI~(DPAI) distributes the perception, the computing and the response modules across a wide space, such as a factory or a city, as shown in Fig.~\ref{fig:dpai}. Industrial~IoT system is a good illustration of DPAI~\cite{iiot}.

\begin{figure}[ht]
\centering
\includegraphics[width=1.0\columnwidth]{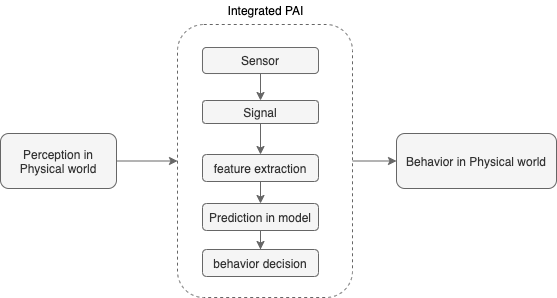}
\caption{Integrated PAI}
\label{fig:ipai}
\end{figure}

\begin{figure*}[ht]
\centering
\includegraphics[width=2.0\columnwidth]{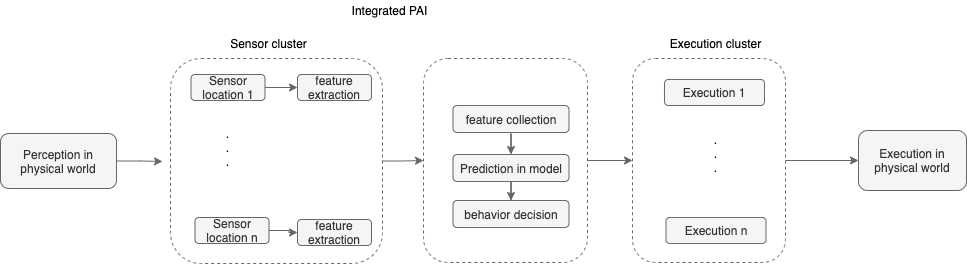}
\caption{Distributed PAI}
\label{fig:dpai}
\end{figure*}

PAI needs to fuse multiple streams of information including materials, temperature, vision, sound,etc. from multiple sensors as per Fig.~\ref{fig:pai}. Therefore, multimodal processing is mandatory to understand the information in PAI. Through the fuse of the multimodal information, PAI can easier use more kinds of information to make better decision and better precisions~\cite{m1,m2}. The data and information sources bring multiple kinds of data, which outperform a single source of data, to make real-time decisions and predictions. This is a significant feature of~PAI.

We use Fig.~\ref{fig:paitwo} to illustrate the components and relations of PAI:IPAI and DPAI. IPAI will be researched and applied in both home environment and industry environment. The home environment~\cite{homer} will receive home service robots like household robots, while the industry environment will be extensively used in multiple areas of the Industry~4.0~\cite{i40} from the automative to the security. DPAI will become more and more popular when the edge computing~\cite{edge} is mature and every device is connected to the network. IoT and edge computing are typical DPAI subdomains. Since it is popular for every intelligent system to be online, IPAI and DPAI will have more overlapped areas as shown in Fig.~\ref{fig:paitwo}.

\begin{figure}[ht]
\centering
\includegraphics[width=1.0\columnwidth]{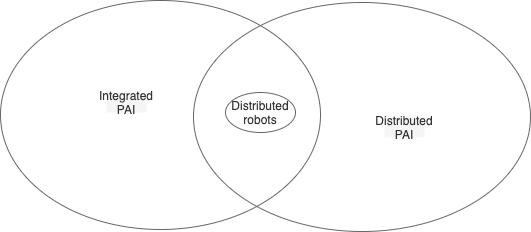}
\caption{IPAI and DPAI}
\label{fig:paitwo}
\end{figure}

\section{The DIKW Supported Physical Artificial Intelligence}

Artificial Intelligence needs a large volume of data as the "fuel" to train the model for the tasks of the classifications and the predictions. Digital Artificial Intelligence such as image classification and automatic speech recognition is typically the approach of processing the signal and data from the sources of the image, the sound, the text and the temporal data. In order to organize the data used in Digital Artificial Intellgence well, the researchers and industry use the Knowledge Graph \cite{kg} to store the ontology from different data. Knowledge graph is a complete and correct approach to associate the semantic data. Kowledge Graph considers all the data inside as the same hierarchical layer, but it does not work very well in the real world. For example, the sentence "the spoiled food can not be eaten" represents one knowledge or a rule, not only the data indicating "food". So DIKW\cite{dikw} architecture is proposed to construct the information architecture. The DIKW architecture is illustrated in Figure \ref{fig:dikw}. In DIKW architecture, the $data$ and $information$ could be used to infer the $knowledge$, while the $wisdom$ as partial $knowledge$ needs the support from the $data$ and $information$. One important feature of DIKW architecture is the presentation the 5 Ws: $Who, What, When, Where$, and $Why$. $Knowledge$ can well describe $What$ happens. $Wisdom$ in DIKW represents $how$. $Data$ is related to $Who$, $When$, and $Where$. And $What$ and $How$ can be infered from $Information$ and $Knowledge$ too.

\begin{figure}[ht]
\centering
\includegraphics[width=0.6\columnwidth]{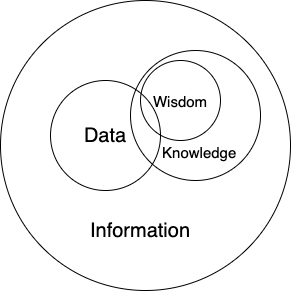}
\caption{DIKW architecture}
\label{fig:dikw}
\end{figure}

Digital Artificial Intelligence originates from data and signal processing, especially the text, image and acoustic processing. In the DIKW architecture, most algorithms of above categories belong to the $Data$ layer. For example, the image object recognition \cite{obj} is to use large volume of object image data to train a model and then recognize an object name in testing images. While, automatic speech recognition is to convert the speech in the sound to the data of the text. In the research, the knowledge extraction in Digital Artificial Intelligence exists but is not as popular as the data extraction. In the article \cite{itot} the authors use the multimodal data processing to extract the knowledge of the image or video, like "One bird flies in the sky". According to the best of our knowledge, it is rare to find extensive deep learning model to deal with more advanced knowledge processing. Therefore, Physical Artificial Intelligence encounters challenging problems as it needs to process the data, information and knowledge and it is not limited to signals as Digital Artificial Intelligence is. PAI needs to accept and process the signal and data from at least five categories: materials, mechanics, chemistry, bilogy, and computer sensors. In order to deal with more categories of signal and data, PAI has to use knowledge graph to support the processing and storage, as illustrated in Figure \ref{fig:graphpai}.

\begin{figure}[ht]
\centering
\includegraphics[width=1.0\columnwidth]{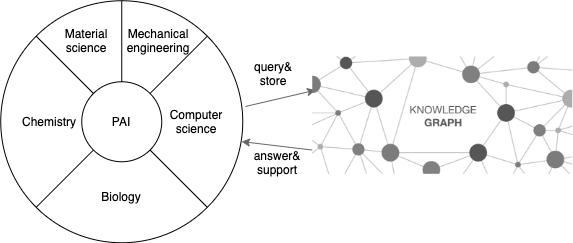}
\caption{Knowledge graph supported PAI}
\label{fig:graphpai}
\end{figure}


As shown in Figure \ref{fig:graphpai}, each node of knowledge graph will contain 5 categories of data from PAI. All data nodes of the same category are internally and organically associated to one another. Knowledge graph could handle the complexity of multiple-indices data. So in Figure \ref{fig:dikwpai} we propose to integrate 5 categoris of data with 5 $W$s and 4 layers of DIKW. Thus the semantic information of PAI could be inferred and stored in DIKW architecture while keep its original relations to the metadata and other basic data.

\section{Physical Artificial Intelligence Governance and Sustainable Development}

Digitial Artificial Intellgience has been facing the challenges of risk and governance problems \cite{gov}. Among the challenges for DIAI, the most important challenges will be discussed in this section:

\begin{itemize}
  \item The security of DIAI \cite{seu}. The training and prediction of AI model needs large volume of data, so the security of the data storage is important. The storage security will need both hardware and software protection. The data masking \cite{mask} is often used to separate the data with the original source in the software and algorithm level protection.
  \item The fake data of DIAI. Deepfake \cite{fake} attracted much attention when it appeared on the Internet. Deepfake could convert the human face in the video to the desired face, and in many situations the coverted video looks real. The fake image and video cause the doubt of "seeing is believing", which could lead to the social and legal problem.
  \item The social privacy of DIAI. The face recognition in the public space has been banned and identified as illegal in many countries \cite{deb}. DIAI has enabled the tracking of our behavior as easy as possible. In addition, DIAI could easily track the online data including the social media and infer the profiles of a person. Therefore, the social privacy has been a big focus in the past years.
  \item The bias in DIAI. In our society the bias exists even if it is hidden, for example the data from the Internet. Most of training data of DIAI is from web source, which means that the training model of DIAI naturally contains the property of bias. This bias has been found in the hiring screening AI system \cite{hiring}.
\end{itemize}

Physical Artificial Intelligence (PAI) has more problems to resolve because of its characteristics of complexity and ubiquitousness compared to DIAI:
\begin{itemize}
  \item The existence problem. PAI like IoT needs more extensive installation of multiple kinds of sensors. If it is in a limited space like a factory, it does not have much regulation problem. However, if the space is extended to a larger space which is not under the same regulation, PAI will face more problems of regulation and social problem.
  \item The information organization problem. As discussed in the previous section, the organization of multiple kinds and multiple layers of data and information will cause the problem of complexity. The proposed Knowledge graph and DIKW supported PAI could be the potential solution.
  \item Cannikin Law. The development of PAI depends on at least 5 disciplines of materials science, mechanical engineering, chemistry, biology and computer science. Therefore, the slower development of one discipline will cause the problem of cannikin law and prohibit the development of PAI.
  \item The social acceptance. Similar to the dilemma of DIAI, the ubiquitous application of PAI will cause the worry of the society regarding to the unemployment, the privacy and etc.
\end{itemize}

We illustrates above problem of PAI in Figure \ref{fig:dikwpai}.  
\begin{figure}[ht]
\centering
\includegraphics[width=0.9\columnwidth]{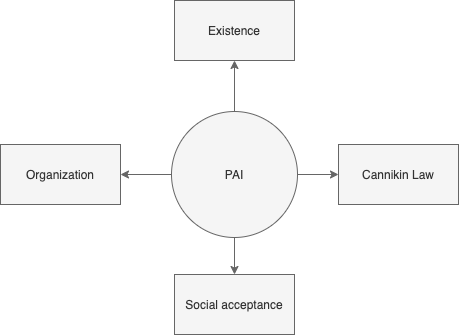}
\caption{PAI governence problems}
\label{fig:dikwpai}
\end{figure}

As the future format of Artificial Intelligence, Physical Artificial Intelligence will be the next popular research topic following the Digital Artificial Intelligence, because Artificial Intelligence will be more and more  applied in other industries. Physical Artificial Intelligence will support the development of the mechanics or the agriculture because of their hardware characteristics. Physical Artificial Intelligence will advance AI application as a fundamental technology for the world.

\section{Conclusion}
In this paper we have started by reviewing the basic knowledge of artificial intelligence, including its history, categories and popular algorithms. Then we reviewed the concept of Physical Artifical Intelligence proposed by Aslan Miriyev and Mirko Kovac, and discussed the reason of extending the concept of Physical Aritificial Intelligence by Integrated Physical Artificial Intelligence and Distributed Physical Artificial Intelligence. After that, we proposed to use DIKW and knowledge graph to extend the concept of Physical Artificial Intelligence. Finally we discussed the governance of Physical Artificial Intelligence and its sustainable development, compared to the current popular topics of Digital Artificial Intelligence governance. We wish to use this paper to discuss the potential development of Physical Artificial Intelligence as the next generation of Artificial Intelligence, and inspire more research and application of Physcial Artifical Intelligence with the discussed theoretical support.


\ifCLASSOPTIONcompsoc
  \section*{Acknowledgments}
\else
\fi

Supported by Natural Science Foundation of China Project (No. 61662021 and No.72062015).



%

\end{document}